\documentclass[10pt,twocolumn,letterpaper]{article}

\usepackage{cvpr}              

\definecolor{cvprblue}{rgb}{0.21,0.49,0.74}
\usepackage[pagebackref,breaklinks,colorlinks,allcolors=cvprblue]{hyperref}
\usepackage{booktabs,multirow,tabularx,makecell}
\usepackage{siunitx}
\usepackage{tikz}
\usepackage{times}
\usepackage{epsfig}
\usepackage{float}
\usetikzlibrary{calc}

\newcolumntype{Y}{>{\raggedright\arraybackslash}X}
\newcommand{\best}[1]{\textbf{#1}}
\newcommand{\second}[1]{\underline{#1}}

\def\paperID{43536} 
\def\confName{CVPR}
\def\confYear{2026}
\renewcommand{\thefootnote}{\fnsymbol{footnote}}

\DeclareMathOperator*{\argmin}{argmin} 
\title{PTC-Depth: Pose-Refined Monocular Depth Estimation with Temporal Consistency}
\author{Leezy Han$^1$, Seunggyu Kim$^1$, Dongseok Shim$^2$, Hyeonbeom Lee$^1$\footnotemark[1]\\
$^1$Ajou University \quad $^2$Sony Creative AI Lab\\
}

\begin{document}
\twocolumn[{%
\renewcommand\twocolumn[1][]{#1}%
\maketitle

\vspace{-1.5em}
\begin{center}
    \centering
    \captionsetup{type=figure}
  \includegraphics[width=\textwidth]{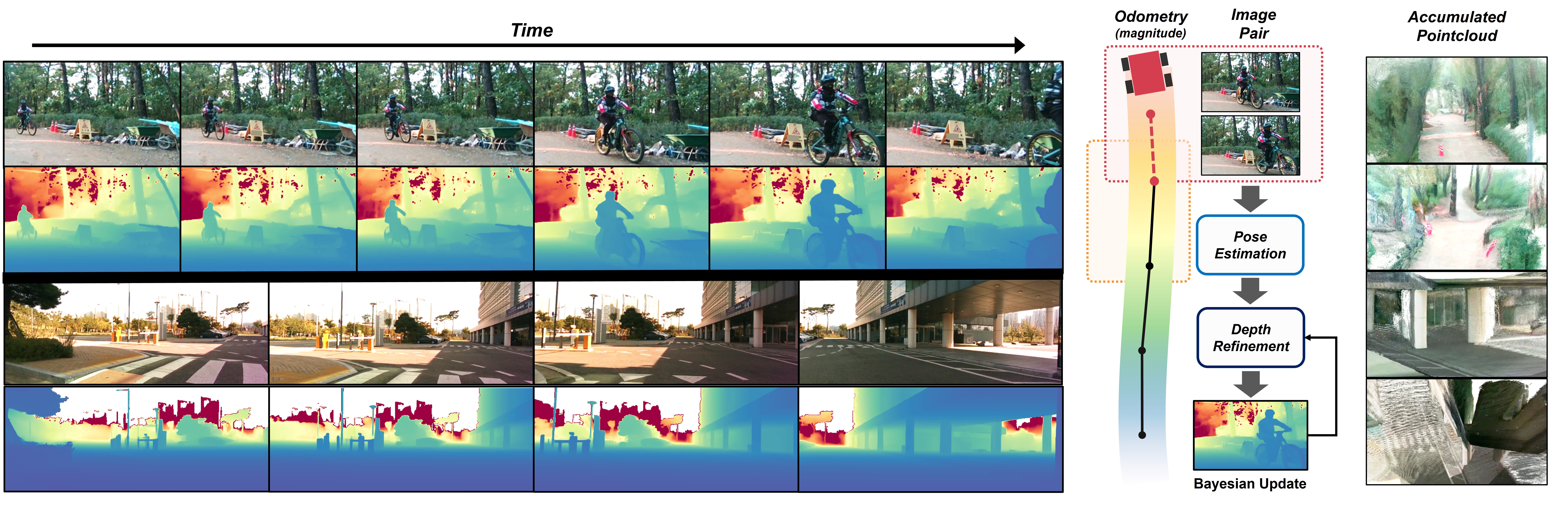}
  \caption{\textrm{\small \textbf{PTC-Depth: metric-consistent monocular depth estimation.} By fusing optical flow with metric displacement from wheel odometry or GPS, our method recovers metric scale and maintains temporal consistency across consecutive frames.}}
  \label{fig:teaser}
\end{center}%
}]

\footnotetext[1]{Corresponding author: hbeomlee@ajou.ac.kr}
\renewcommand{\thefootnote}{\arabic{footnote}}

\begin{abstract}
Monocular depth estimation (MDE) has been widely adopted in the perception systems of autonomous vehicles and mobile robots. However, existing approaches often struggle to maintain temporal consistency in depth estimation across consecutive frames. This inconsistency not only causes jitter but can also lead to estimation failures when the depth range changes abruptly. To address these challenges, this paper proposes a consistency-aware monocular depth estimation framework that leverages wheel odometry from a mobile robot to achieve stable and coherent depth predictions over time. Specifically, we estimate camera pose and sparse depth from triangulation using optical flow between consecutive frames. The sparse depth estimates are used to update a recursive Bayesian estimate of the metric scale, which is then applied to rescale the relative depth predicted by a pre-trained depth estimation foundation model. The proposed method is evaluated on the KITTI, TartanAir, MS2, and our own dataset, demonstrating robust and accurate depth estimation performance. The project page is available at \href{https://ptc-depth.github.io}{https://ptc-depth.github.io}.
\end{abstract}    
\section{Introduction}

Deep learning–based depth estimation, which relies solely on a monocular camera, has recently been applied to a diverse range of tasks, including surgical scene understanding \cite{surgical}, 3D scene reconstruction for autonomous vehicles \cite{zheng2024physical,parking}, and robotic perception \cite{arrltie, wu2022toward,UDepth,r360monodepth}. To enable its broader applicability, recent research has focused on improving the accuracy and robustness of depth estimation techniques. With the emergence of foundation models for depth estimation trained on large-scale datasets \cite{depthanyv2, depthpro, marigold, unidepth}, the performance of monocular depth prediction has steadily advanced. Several of these models leverage synthetic data generation to construct large-scale training datasets, and employ zero-shot generalization to achieve strong performance across diverse environments. Nevertheless, maintaining temporal consistency across consecutive frames remains a major challenge for real-world applications.

Several approaches have been proposed to address the problem of inconsistent depth estimation across consecutive frames, including single-image relative depth estimation methods \cite{midas} extended to video sequences \cite{rollingdepth} and depth compensation techniques that incorporate sparse depth measurements \cite{marigolddc, lin2025prompting}. However, in video-based learning, the absence of absolute distance information limits its applicability to autonomous vehicles and robotic systems. Furthermore, methods such as those in \cite{marigolddc, lin2025prompting} require additional sensors and rely on metric depth images, making them incompatible with foundation models that provide only relative depth predictions.

In this paper, we propose a depth estimation framework that enhances the temporal consistency of monocular depth predictions without relying on additional sparse depth data or depth sensors. Our contributions are:
\begin{itemize}
\item Our method produces temporally consistent metric depth maps by tracking the \textit{metric scale} of a foundation depth model's relative output via recursive Bayesian updates. The key insight is that wheel odometry and optical flow directly constrain scale, and aggregating per-pixel estimates via superpixels enables robust Bayesian scale tracking even under odometry noise.

\item By leveraging a foundation depth model, our method achieves metric depth estimation while preserving sharp boundary quality across diverse imaging modalities, including RGB and long-wave infrared (LWIR) images.
\item Our method achieves temporally consistent metric depth with higher accuracy than existing video-based depth methods \cite{rollingdepth,chen2025video}, without relying on additional depth sensors or pre-trained metric priors.
\item To evaluate the performance of our algorithm, we conducted experiments on several benchmark datasets, including RGB datasets such as KITTI\cite{kitti}, TartanAir\cite{tartanair}, and a thermal dataset using MS2\cite{shin2023deep}. We also collected our own dataset in diverse environments, including forest and urban areas, for additional evaluation. 
\end{itemize}

\section{Related Works}

\subsection{Monocular Depth Estimation}
Deep learning–based depth estimation has motivated the development of self-supervised approaches \cite{monodepth2,erdepth,zhang2022tim,zhang2022iros,swindepth,lavreniuk2025spidepth, guizilini2022learning}. Self-supervised monocular depth estimation methods jointly train a DepthNet and a PoseNet, where the PoseNet predicts the relative camera motion between frames to enforce photometric consistency. However, these self-supervised methods often suffer from performance degradation under illumination changes and in the presence of moving objects, leading to limited generalization capability.

Metric depth estimation from a single RGB image \cite{adabins,yuan2022new, depthpro,metric3d,sharpdepth,shim2024sediff,bhat2023zoedepth} achieves state-of-the-art performance through large-scale supervised training. However, these approaches often exhibit limited generalization capability, showing blurred boundaries and degraded performance when applied to environments different from the training environment. Recent foundation model–based methods \cite{depthanyv2, depthpro, marigold, unidepth} have explored learning frameworks that combine both metric depth and relative depth, where the latter captures per-pixel depth ratios without predicting absolute metric scale. Unlike metric depth prediction, relative depth estimation focuses on learning the geometric relationships between pixels rather than predicting absolute metric scales tied to specific camera intrinsics or scene configurations. Relative depth models are known to achieve superior zero-shot generalization performance across diverse domains and imaging conditions. DepthFM \cite{gui2025depthfm} further explores generative approaches by formulating depth estimation as a flow matching problem, enabling fast inference through straight ODE trajectories.

Due to these challenges, the development of video-based depth estimation models \cite{zhang2019exploiting,kopf2021robust,rollingdepth,chen2025video,wang2023neural,zhu2023lighteddepth} aimed at improving temporal consistency has been actively studied. Most of these approaches estimate relative depth, which contributes to smoother temporal transitions and improved frame-to-frame coherence compared to conventional monocular camera–based methods. Since these depth models do not provide absolute metric information, they are difficult to apply to tasks such as autonomous driving or 3D mapping, where accurate scale estimation is essential. Some methods have attempted to address this by jointly estimating depth and camera motion \cite{yang2020d3vo}, or by leveraging depth priors to improve camera pose estimation \cite{madpose,ding2025reposed}. However, these approaches focus primarily on pose accuracy and do not address temporal scale consistency in dense depth maps.

\subsection{Depth Completion}

Depth completion techniques produce dense depth maps by fusing RGB images with sparse depth point clouds. Early works relied on classical interpolation applied directly to sparse depth inputs \cite{chodosh2018deep} or used RGB images as guidance \cite{ma2018sparse}. With advances in deep learning, more accurate depth completion methods have emerged, including stereo-based approaches \cite{bartolomei2024revisiting} and diffusion-based methods that interpolate sparse depth using RGB information \cite{liu2024depthlab}. These approaches exhibit satisfactory performance across a wide range of scenarios. However, diffusion-based approaches such as \cite{liu2024depthlab} rely solely on relative depth and cannot provide absolute metric depth information.

While video-based depth estimation seeks to obtain temporally consistent depth from multi-frame inputs using relative depth, recent depth completion approaches \cite{videpth,marigolddc,lin2025prompting,marsal2024simple} aim to estimate highly reliable metric depth by combining reliable monocular depth with pre-provided sparse depth measurements. However, these methods require additional depth sensors such as LiDAR, making them impractical when only a camera and odometry are available. Our method eliminates this requirement, relying solely on wheel odometry as a lightweight metric reference.

\begin{figure*}[t]  
  \centering
  \includegraphics[width=\linewidth]{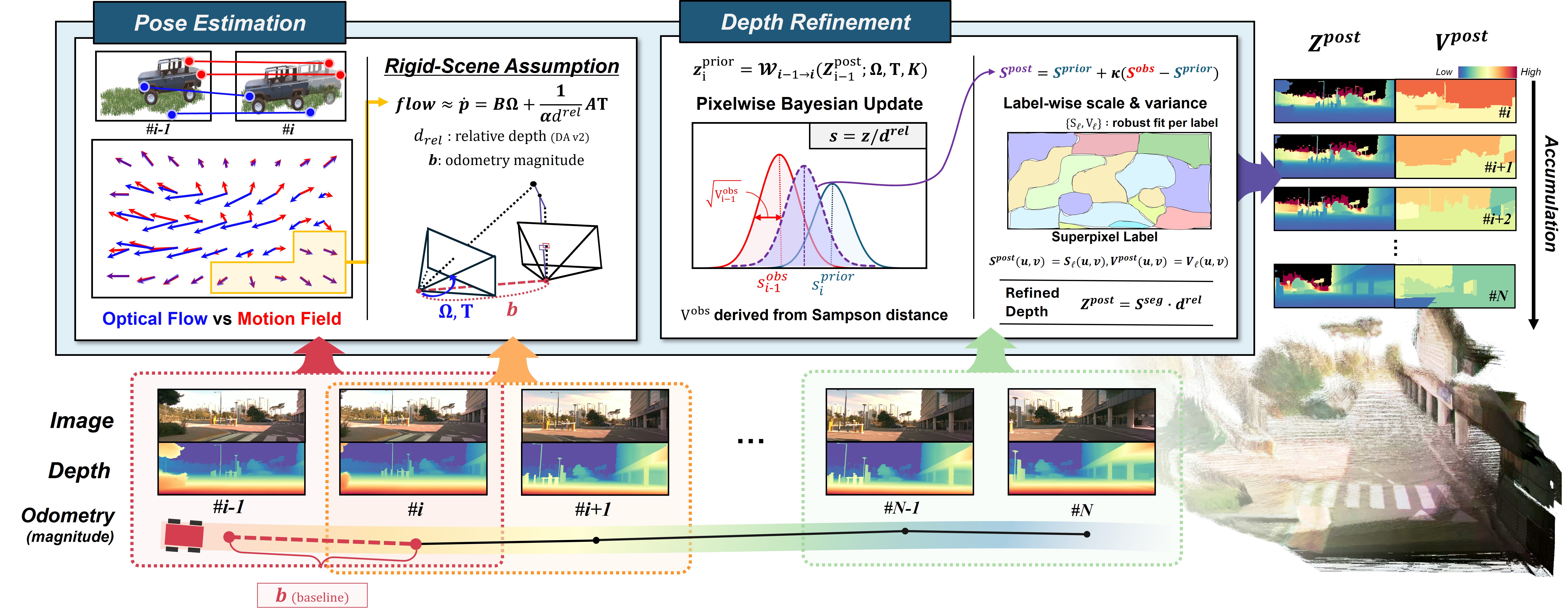}
  \caption{\textrm{\small \textbf{The overall framework of the proposed method.} Camera pose and metric scale are estimated from optical flow and wheel odometry, respectively, and used to fuse triangulated metric depth with relative depth from a foundation model.}}
  \label{fig:framework}
\end{figure*}

\section{Method}
The overall framework of the proposed method is shown in Fig. \ref{fig:framework}. Our approach first estimates the optical flow between two consecutive image frames, from which camera pose and metric scale are recovered using wheel odometry. Subsequently, a sparse depth map is derived from the estimated pose and optical flow information. Finally, the sparse depth is fused with the relative depth predicted by a foundation model \cite{depthanyv2} through recursive Bayesian updates, resulting in a consistent depth map that incorporates accurate metric scale information.

\subsection{Observed Flow and Motion Field}
\label{sec:mf}
Following the classical Longuet-Higgins–Prazdny formulation~\cite{Heeger1992SubspaceMF}, camera motion induces a motion field on the image plane. The components of this motion field in normalized camera coordinates $(x,y)$ can be expressed in terms of the rotation $\boldsymbol{\Omega}$ and translation ${\boldsymbol{T}}$ of the camera between frames as follows:
\begin{align}
\dot{\mathbf{p}}(x,y)=\mathbf{B}(x,y)\,\boldsymbol{\Omega}
+\frac{1}{\alpha d^\text{rel}}\,\mathbf{A}(x,y)\,\boldsymbol{T},
\label{eq:mf}
\end{align}
where $d^\text{rel}$ denotes the relative depth predicted by an affine-invariant monocular depth model. Note that $d^\text{rel}$ is obtained by inverting the raw inverse-depth output of the monocular network. $\alpha$ is a single global scale factor that converts the relative depth into metric scale. For simplicity, we assume that the relative depth $d^\text{rel}$ can be converted into the metric depth using a single scale parameter $\alpha$. Although this assumption seems stronger than those employed in scale-and-shift depth compensation methods, we adopt it for the sake of simplicity. Since the depth refinement process divides the image into superpixels and assigns an individual scale to each cell, this assumption does not impose a strong limitation on the scale-and-shift problem.
The matrices $\mathbf{A}(x,y)$ and $\mathbf{B}(x,y)$ in \eqref{eq:mf} are defined as
\begin{align}
\mathbf{A}=
\begin{bmatrix}
-1 & 0 & x\\[2pt]
0 & -1 & y
\end{bmatrix},\
\mathbf{B}=
\begin{bmatrix}
xy & -(1+x^2) & y\\[2pt]
1+y^2 & -xy & -x
\end{bmatrix}. \nonumber
\end{align}

Since the motion observed between consecutive image frames is closely related to the actual movement of the mobile robot, we can reformulate the translation vector ${\boldsymbol{T}}$ as $\boldsymbol{T} = b\,\hat{\boldsymbol{T}}$, where $b$ denotes the odometry-derived baseline and $\hat{\boldsymbol{T}}$ represents the normalized direction vector. Given $N$ flow measurements indexed by $n = 1, \ldots, N$, the resulting motion constraints can be formulated as:
\begin{align}
\begin{bmatrix}
\dot{\mathbf{p}}_1 \\ \vdots \\ \dot{\mathbf{p}}_N
\end{bmatrix}
\approx
\begin{bmatrix}
\mathbf{B}_1 & \frac{b}{d^{\text{rel}}_1}\,\mathbf{A}_1 \\
\vdots & \vdots \\
\mathbf{B}_N & \frac{b}{d^{\text{rel}}_N}\,\mathbf{A}_N
\end{bmatrix}
\begin{bmatrix}
\boldsymbol{\Omega} \\[2pt]
\frac{\boldsymbol{\hat{T}}}{\alpha}
\end{bmatrix}.
\label{eq:blocks}
\end{align}
To recover $[\boldsymbol{\Omega}, \boldsymbol{T}]^T$, we normalize the term $\boldsymbol{\hat{T}}/\alpha$, i.e., $\frac{\hat{\boldsymbol{T}}}{\alpha}/\|\frac{\hat{\boldsymbol{T}}}{\alpha}\| = \hat{\boldsymbol{T}}$, given that $\|\hat{\boldsymbol{T}}\| =1$. The full translation $\boldsymbol{{T}}$ can then be reconstructed as $\boldsymbol{T} = b\,\hat{\boldsymbol{T}}$. The estimation of scale factor $\alpha$ will be described in a later section. To compute optical flow, we use DIS Flow~\cite{disflow} for RGB images. For thermal images, FieldScale~\cite{Fieldscale} is applied as preprocessing prior to optical flow estimation.

\subsection{Robust Pose Estimation and Scale Recovery}

When estimating $\boldsymbol{\Omega}$ and $\boldsymbol{T}$ from \eqref{eq:blocks}, two major issues must be addressed. 

\noindent\textbf{Dynamic-object outliers} Flow measurements from independently moving objects do not reflect camera motion and must be excluded (Fig.~\ref{fig:framework}). However, because the motion field depends on the unknown parameters $(\boldsymbol{\Omega}, \boldsymbol{T})$, such outliers cannot be identified without first knowing the motion parameters. For this reason, we employ RANSAC to obtain a robust estimate of the camera motion.

We compute the residual between the observed optical flow and the predicted motion field. Since flow magnitude varies with depth, the residual is normalized by the flow magnitude to maintain a consistent inlier criterion across the image. Flow direction also provides a useful cue: when a flow vector deviates significantly in direction from the predicted motion, it is classified as an outlier. If the matched flow vectors originate primarily from a small spatial region, the estimated motion may fail to represent the full scene. To mitigate this bias, we partition the image into grid cells and sample an equal number of flow vectors from each cell. RANSAC hypotheses with inliers covering a larger number of cells are preferred, ensuring that the final motion estimate is valid across the entire field of view. The best-performing RANSAC hypothesis is further refined using IRLS with Huber weighting as described in \cite{hartley2003multiple}. The detailed equations and procedures are provided in the supplementary material.

\noindent\textbf{Scale ambiguity of translation}
The translation $\boldsymbol{T}$ is determined only up to an unknown scale factor, so external information is required to recover its true magnitude. The baseline ${b}$ is obtained from wheel odometry or other sources such as GPS. However, these external sensors are not perfectly synchronized with the RGB/thermal images, so the recovered scale can still be inaccurate. The resulting scale error manifests as noise in the triangulated depth observations and is handled by the Bayesian fusion described in the next section.

\subsection{Triangulation and Sampson Residual}
\label{sec:tri}
Given the relative pose $(\mathbf{\Omega}, \mathbf{T})$ estimated from the motion field, we compute the metric depth at frame $i$ by intersecting the viewing rays of a corresponding pixel pair 
$(u_{i-1}, v_{i-1}) \leftrightarrow (u_i, v_i)$, where 
$\mathbf{x}_i = [u_i, v_i, 1]^\top$ and 
$\bar{\mathbf{x}}_i = K^{-1}\mathbf{x}_i$ denotes its normalized camera coordinate with $K$ the intrinsic matrix.
We find $(z_{i-1},\, z_i)$ minimizing the residual between points on the two viewing rays:
\begin{equation}
    (z^{\mathrm{tri}}_{i-1},\, z^{\mathrm{tri}}_i)
    = \argmin_{z_{i-1},\, z_i}
    \left\|z_{i-1}\,\mathbf{\Omega}\,\bar{\mathbf{x}}_{i-1} + \mathbf{T}
    - z_i\,\bar{\mathbf{x}}_{i}\right\|.
\end{equation}
Here, the minimizer $z^{\mathrm{tri}}_i$ gives the triangulated metric depth at frame $i$. These per-pixel values are assembled into the sparse depth map $z^{\mathrm{tri}}_i \rightarrow \bold{Z}^{\mathrm{tri}}_i$, used as the observation in the Bayesian fusion (Section~\ref{sec:fusion}). The triangulated observation and the temporal prior warped from the previous frame are fused to produce the posterior depth, which is then recursively propagated to the next frame.

\noindent\textbf{Sampson Residual} Triangulation can fail due to flow errors,
moving objects, or pose inaccuracies. Since reprojection error requires known
depth a priori, we instead use the Sampson residual~\cite{hartley2003multiple},
which evaluates how well each correspondence satisfies the epipolar constraint.
We treat this as a per-pixel geometric reliability score: small values indicate
epipolar-consistent matches, while large values flag unreliable triangulations.
We assign each Sampson residual to its corresponding pixel in frame $i$ to form a \textit{Sampson residual map}. It is used both to weight each triangulated depth observation and to adaptively inflate the prior variance before fusion. The detailed computation of the \textit{Sampson residual map} is described in the supplementary material. 

\begin{figure}
  \centering
  \includegraphics[width=\columnwidth]{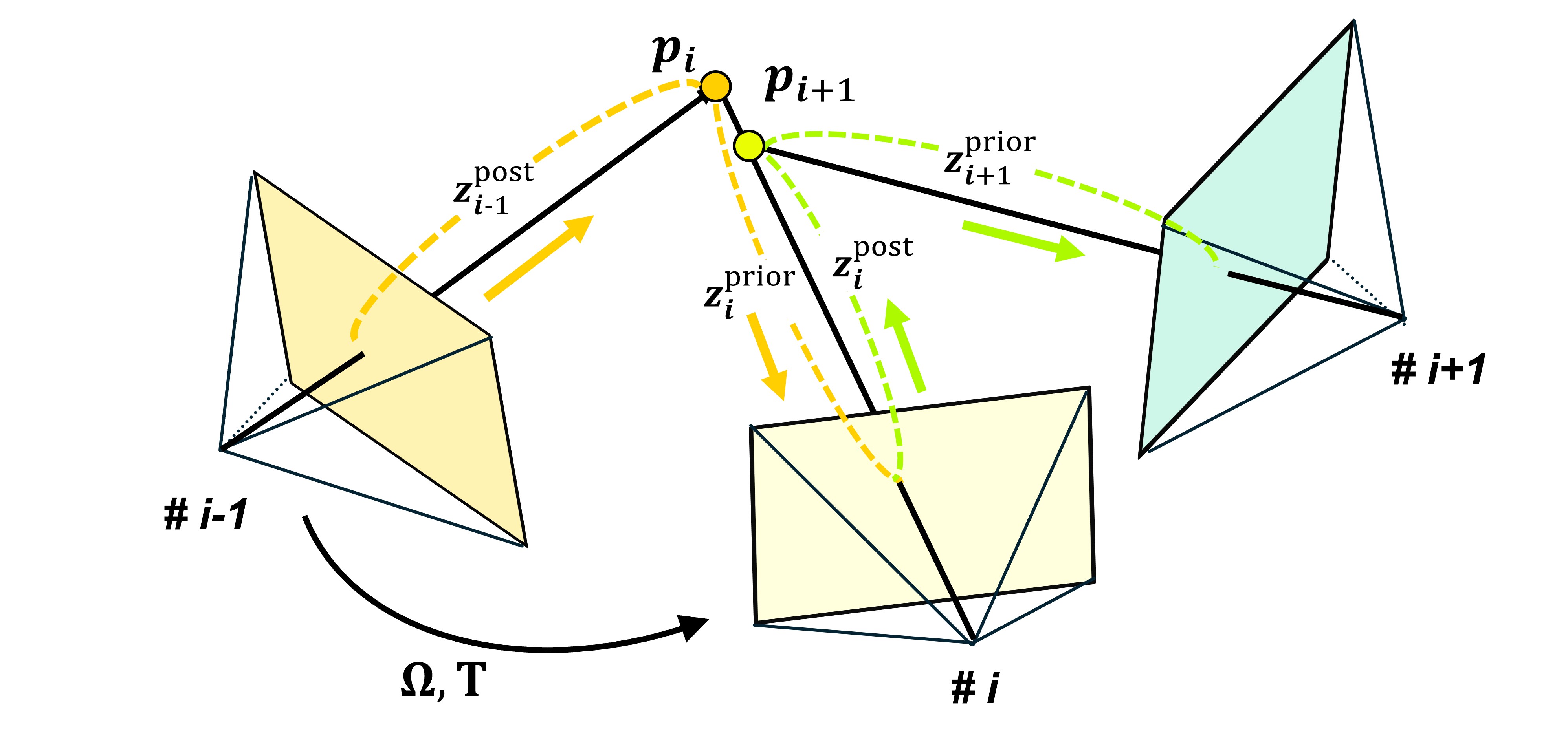}
  \caption{\textrm{\small \textbf{Triangulation and temporal depth propagation.}
  The posterior depth $z^{\mathrm{post}}_{i-1}$ from frame $i{-}1$ is
  lifted to the 3D point $\mathbf{p}_i$, transformed by the estimated
  pose $(\boldsymbol{\Omega},\boldsymbol{T})$, and projected into
  frame $i$ to form the temporal prior $z^{\mathrm{prior}}_i$.
  Independently, triangulation from optical flow provides a new
  metric observation. Bayesian fusion combines the temporal prior
  with this observation to produce the refined posterior
  $z^{\mathrm{post}}_i$, which is recursively propagated to the next
  frame as $z^{\mathrm{prior}}_{i+1}$.}}
  \label{fig:zpost}
\end{figure}

\noindent \textbf{Warping Previous Depth} As shown in Fig. \ref{fig:zpost}, to obtain a depth prior for the current frame, we warp the posterior depth map $\bold{Z}^\text{post}_{i-1}$, which represents the final metric depth inferred for the previous frame after Bayesian fusion (Section~\ref{sec:fusion}), into frame $i$ using the estimated relative pose $(\mathbf{\Omega}, \mathbf{T})$. Each pixel $\mathbf{x}_{i-1}$ is converted to a 3D point
\begin{equation}
    \mathbf{p}_{i-1} = z^\text{post}_{i-1}\,\mathbf{\bar{x}}_{i-1}, \label{eqn:pi-1}
\end{equation}
and transformed into frame $i$ via the estimated pose:
\begin{align}
    \mathbf{p}_{i} = \mathbf{\Omega}\,\mathbf{p}_{i-1} + \mathbf{T}, \qquad
    \mathbf{x}_{i} = \mathrm{proj}(K\,\mathbf{p}_{i}). \label{eqn:zpior}
\end{align}
$z^\text{prior}_i$ is the depth at $\bold{x}_i$ and can be computed from $\mathbf{p}_{i}$. For clarity in the subsequent exposition, we define the overall process described in \eqref{eqn:pi-1}-\eqref{eqn:zpior} using the operator $\mathcal{W}$ as follows:
\begin{equation}
\label{eq:warp}
    \bold{Z}^\text{prior}_{i} = \mathcal{W}_{i-1 \rightarrow i}(\bold{Z}^\text{post}_{i-1};\, \mathbf{\Omega},\, \mathbf{T},\, K).
\end{equation}
In the following section, we describe how to estimate the true depth by fusing $\bold{Z}^\mathrm{tri}_i$ and $\bold{Z}^\text{prior}_{i}$ through Bayesian fusion.
 
\subsection{Bayesian Scale Fusion}
\label{sec:fusion}

Triangulation provides metrically meaningful depth but suffers from noise and instability under short baselines or imperfect correspondence. In contrast, relative depth $d^{\mathrm{rel}}$ offers a smooth and structurally coherent representation but lacks metric scale. Rather than fusing the two directly in depth space, which would discard the structural coherence of $d^{\mathrm{rel}}$, we estimate a latent scale field $S$ such that $\bold{Z} = S \cdot d^{\mathrm{rel}}$ recovers metric depth, along with a per-pixel variance map $V$.

\vspace{-1em} \paragraph{Prior from previous frame} From (\ref{eq:warp}), the prior scale is $S^\mathrm{prior} = Z^\mathrm{prior} / d^\mathrm{rel}$, with $V^\mathrm{post}_{i-1}$ propagated to $V^\mathrm{prior}_i$ via the same warp $\mathcal{W}_{i-1 \rightarrow i}$. We drop the frame index hereafter.

When the overall frame geometry is unreliable, the warp $\mathcal{W}$ itself may be inaccurate, making the propagated prior less trustworthy. We therefore use the median Sampson residual $\tilde{\rho} = \mathrm{median}(\rho)$ over the current frame as a proxy for frame-level geometric quality and inflate the prior uncertainty uniformly:
\begin{align}
  V^{\mathrm{prior}} \leftarrow V^{\mathrm{prior}} 
  \left(1 + \frac{\tilde{\rho}}{f_x f_y}\right),
\end{align}
where $f_x$ and $f_y$ are the focal lengths in pixels, and dividing by $f_x f_y$ converts the pixel-domain residual to an angular quantity invariant to image resolution.
\vspace{-1em} \paragraph{Observation from triangulation}
The triangulated depth $\bold{Z}^{\mathrm{tri}}$ directly induces an observed scale $S^{\mathrm{obs}} = \bold{Z}^{\mathrm{tri}} / d^{\mathrm{rel}}$.
Its per-pixel uncertainty is derived from the Sampson residual 
$\rho$ at each pixel as $V^{\mathrm{obs}} = \sigma^2 \frac{\rho}{f_x f_y}$. Here, $\sigma^2$ is a fixed variance scale.
A larger correspondence error implies a larger angular uncertainty in the viewing ray, which propagates to a proportionally larger uncertainty in the triangulated depth and hence in $S^{\mathrm{obs}}$.

\vspace{-1em} \paragraph{Bayesian update} Given prior $(S^{\mathrm{prior}}, V^{\mathrm{prior}})$ and observation
$(S^{\mathrm{obs}}, V^{\mathrm{obs}})$, we adopt a scalar Bayesian update.
The normalized squared innovation
\begin{align}
\gamma = \frac{(S^{\mathrm{obs}} - S^{\mathrm{prior}})^{2}}
         {V^{\mathrm{prior}} + V^{\mathrm{obs}}}
\end{align}
is tested against the $99\%$ chi-square threshold ($1$ DOF).
Pixels exceeding this threshold retain the lower-variance estimate, preserving the more reliable of the two rather than discarding the update entirely. For inlier pixels, the standard Kalman gain is
\begin{align}
\kappa_{\mathrm{raw}} =
\frac{V^{\mathrm{prior}}}{V^{\mathrm{prior}} + V^{\mathrm{obs}}}.
\label{eqn:gain}
\end{align}
The raw gain in (\ref{eqn:gain}) can be aggressive where triangulation and prior only weakly agree. We therefore cap it using a per-pixel 
consistency score $c \in [0,1]$ computed from the relative discrepancy 
$|S^{\mathrm{obs}} - S^{\mathrm{prior}}| / S^{\mathrm{obs}}$ via a Gaussian-like kernel, where the tolerance is estimated as a median absolute deviation (MAD) over the frame and smoothed with an exponential moving average across frames:
\begin{align}
\kappa = \min\!\left(
\kappa_{\mathrm{raw}},\;
\kappa_{\min} + (1 - \kappa_{\min})\,c
\right),
\end{align}
where $\kappa_{\min}$ is a lower bound ensuring a minimum update strength even in low-consistency regions.
The posterior scale and variance then follow:
\begin{align}
S^{\mathrm{post}} &=
S^{\mathrm{prior}} + \kappa\,(S^{\mathrm{obs}} - S^{\mathrm{prior}}), \nonumber \\
V^{\mathrm{post}} &=
(1-\kappa)^2\,V^{\mathrm{prior}} + \kappa^2\,V^{\mathrm{obs}},
\end{align}
where the Joseph form of the variance update ensures numerical stability for suboptimal gains.

\vspace{-1em} \paragraph{Segment-wise scale consolidation}
As noted in Section~\ref{sec:mf}, the scale recovered from wheel odometry is a multiplicative correction to $d^{\mathrm{rel}}$. A single global scale cannot fully account for the shift component inherent in affine-invariant model outputs. Partitioning the image into superpixels reduces the influence of the shift within each local region, allowing per-segment scale estimation to better capture the true metric depth. We therefore employ Felzenszwalb segmentation~\cite{felzenszwalb2004efficient} to partition the image into superpixels whose boundaries follow the geometry of $d^{\mathrm{rel}}$, and estimate a per-segment scale independently.

Within each segment $\Lambda_\ell$, we compute the median scale $\bar{s}_\ell = \mathrm{median}_{k \in \Lambda_\ell}(S^\mathrm{post}_k)$ over the per-pixel posterior scales $S^\mathrm{post}_k$ and evaluate the fitting error. Segments with sufficient evidence and low fitting error are assigned $S^\mathrm{seg}_k = \bar{s}_\ell$ for each pixel $k \in \Lambda_\ell$. All other segments fall back to a global scale estimate.

This approach is particularly effective in scenes with diverse geometry such as TartanAir, where a single global scale is insufficient.
\vspace{-1em} \paragraph{Final depth}
The final metric depth is obtained as
\begin{align}
    \bold{Z}^{\mathrm{post}} = S^{\mathrm{seg}} \cdot d^{\mathrm{rel}}.
\end{align}
This fusion combines the geometric fidelity of triangulation, the temporal stability of propagated priors, and the structural coherence of segment-wise scale selection, enabling robust and consistent metric depth under flow noise, dynamic objects, and short baselines.

\section{Evaluation}

\begin{figure*}[t]  
  \centering
  \includegraphics[width=\linewidth]{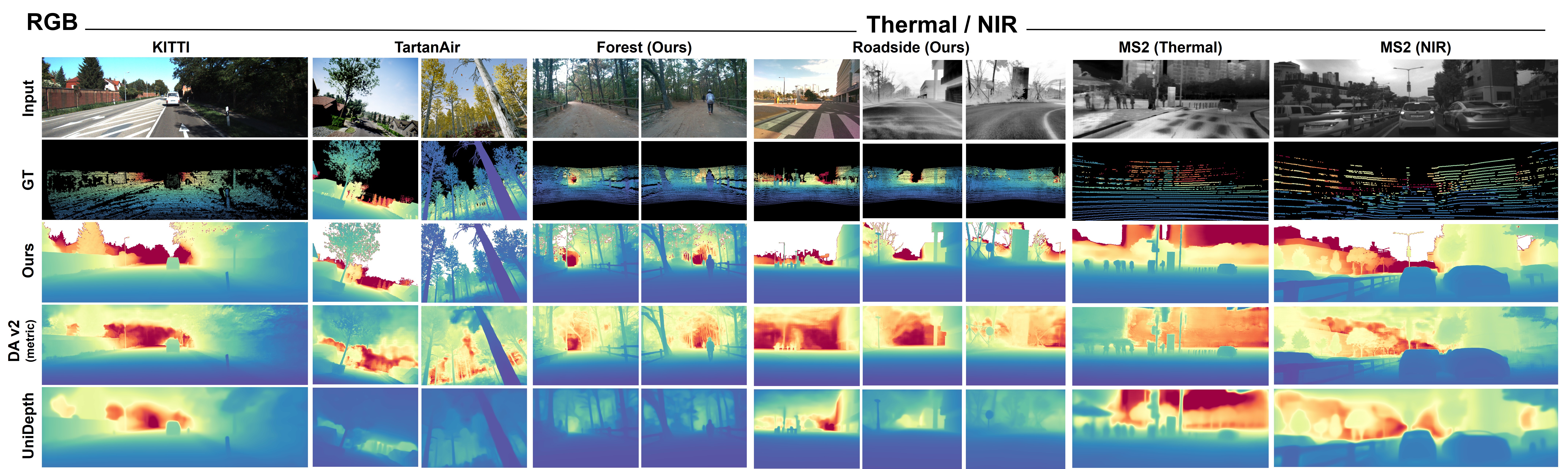}
  \caption{\textrm{\small \textbf{Depth estimation on various datasets.} We evaluated our method on both real-world RGB datasets, such as KITTI, and synthetic RGB datasets, such as TartanAir. We also perform the depth estimation on thermal and NIR imagery using the MS2 dataset as well as our newly collected dataset.}}
  \label{fig:rgbt}
\end{figure*}

\begin{figure*}[t]  
  \centering
  \includegraphics[width=\textwidth]{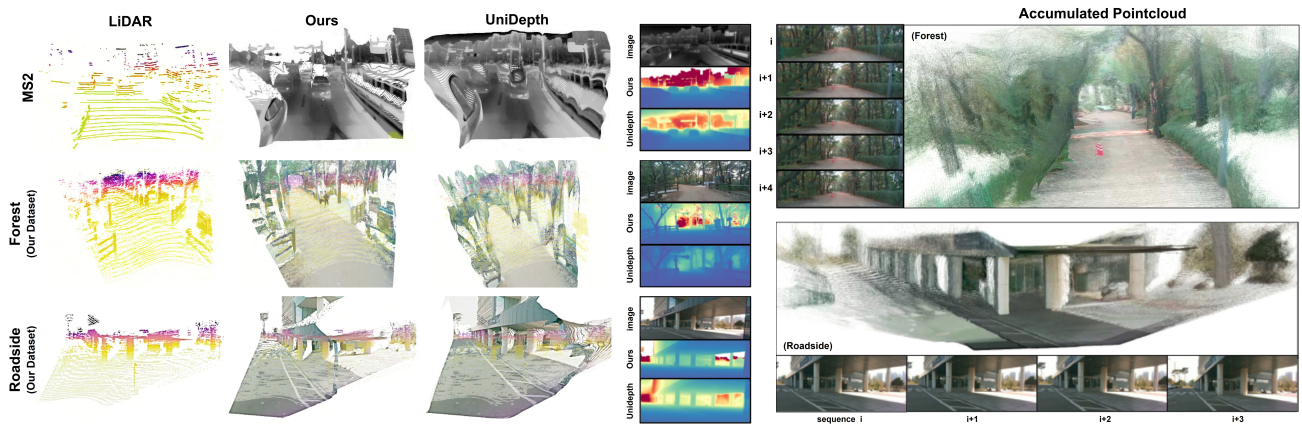}
  \caption{\textrm{\small \textbf{Comparison result of 3D reconstruction.} Compared to conventional depth estimation, our approach produces reliable depth closely aligned to the ground-truth data. Furthermore, when multiple frames are accumulated, the reconstructed point clouds exhibit sufficient geometric consistency.} }
  \label{fig:pt}
\end{figure*}

\begin{figure}
  \centering
  \includegraphics[width=\columnwidth]{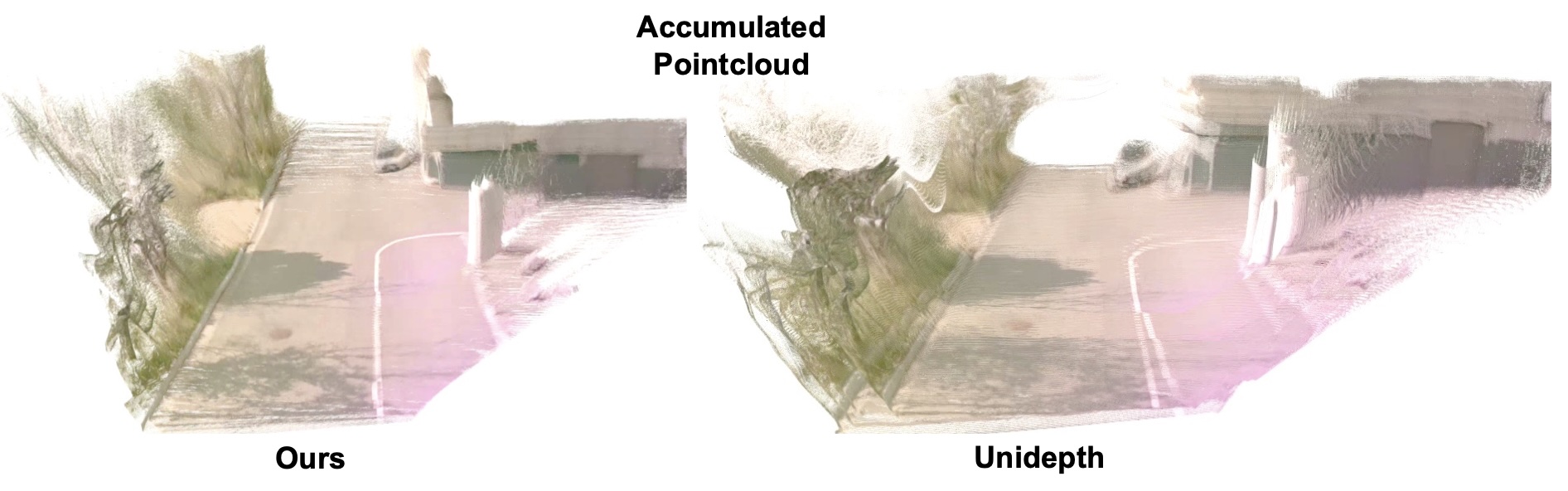}
  \caption{\textrm{\small \textbf{ Comparison result of accumulated pointcloud.} Through the accumulation of multiple point clouds, our algorithm consistently produces temporally coherent depth estimates. } }
  \label{fig:acc}
\end{figure}

\begin{figure}
  \centering
  \includegraphics[width=\columnwidth]{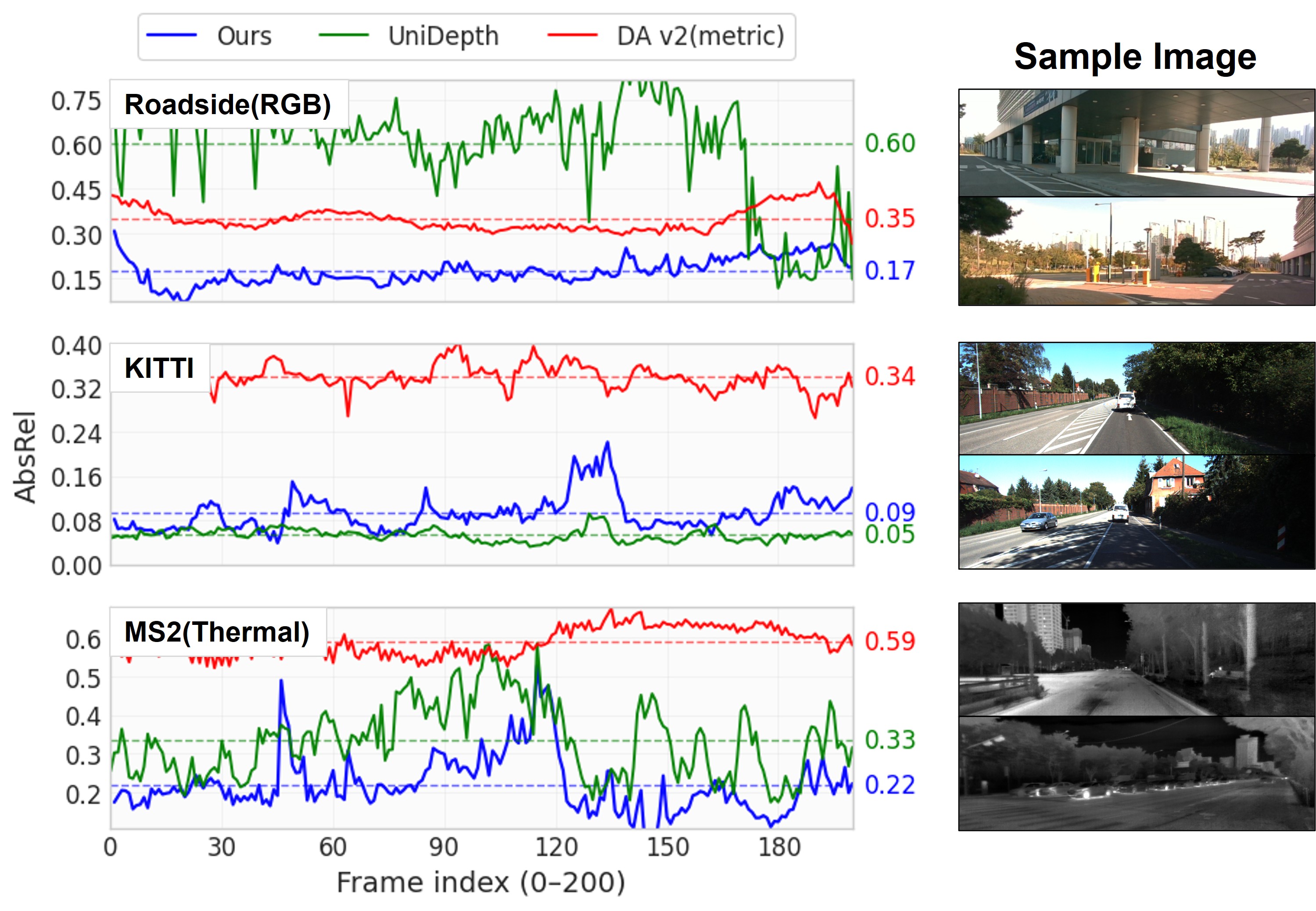}
  \caption{\textrm{\small \textbf{AbsRel error over time.} The line plot shows the depth error at every individual frame, end of each curve indicates the average error across the image sequence.}}
  \label{fig:absrel}
\end{figure}

\begin{table*}[t]
\centering
\resizebox{\textwidth}{!}{
\begin{tabular}{l cc cc cc cc cc cc}
\toprule
& \multicolumn{8}{c}{\textbf{RGB}} &
  \multicolumn{4}{c}{\textbf{Thermal}} \\
\cmidrule(lr){2-9} \cmidrule(lr){10-13}
& \multicolumn{2}{c}{KITTI} &
  \multicolumn{2}{c}{TartanAir} &
  \multicolumn{2}{c}{Roadside} &
  \multicolumn{2}{c}{Forest} &
  \multicolumn{2}{c}{MS2} &
  \multicolumn{2}{c}{Roadside} \\
\cmidrule(lr){2-3} \cmidrule(lr){4-5} \cmidrule(lr){6-7}
\cmidrule(lr){8-9} \cmidrule(lr){10-11} \cmidrule(lr){12-13}
Method
& AbsRel$\downarrow$ & $\delta\!<\!1.25\uparrow$
& AbsRel$\downarrow$ & $\delta\!<\!1.25\uparrow$
& AbsRel$\downarrow$ & $\delta\!<\!1.25\uparrow$
& AbsRel$\downarrow$ & $\delta\!<\!1.25\uparrow$
& AbsRel$\downarrow$ & $\delta\!<\!1.25\uparrow$
& AbsRel$\downarrow$ & $\delta\!<\!1.25\uparrow$ \\
\midrule
MADPose (UniDepth)
& \second{0.115} & \best{0.869}
& 0.481 & 0.187
& 0.423 & 0.222
& 0.473 & 0.097
& \second{1.381} & \second{0.179}
& 0.509 & 0.149 \\
Ours$^{\dag}$ (DA v2 rel)
& \best{0.115} & \second{0.867}
& \second{0.239} & \second{0.645}
& \best{0.227} & \best{0.649}
& \best{0.326} & \best{0.540}
& \best{0.950} & \best{0.332}
& \best{0.346} & \best{0.484} \\
GT Pose
& 0.130 & 0.839
& \best{0.168} & \best{0.754}
& - & -
& - & -
& 1.465 & 0.073
& - & - \\
\bottomrule
\end{tabular}}
\caption{\textbf{Triangulated depth evaluation.} Each method triangulates sparse depth using its own pose source. MADPose relies on a pretrained metric depth model (UniDepth); ours uses odometry only. GT Pose serves as a reference to illustrate the effect of temporal misalignment in real sequences. Errors are robust means over valid pixels (top-90\%).}\label{tab:tri}
\end{table*}

\subsection{Evaluation Datasets}

We evaluate across six scenarios from five datasets without any dataset-specific fine-tuning or retraining. KITTI~\cite{kitti} provides RGB images with sparse LiDAR depth and RTK-GPS odometry. TartanAir~\cite{tartanair} is a photo-realistic synthetic benchmark covering diverse visual conditions. The MS2 dataset~\cite{shin2023deep} provides thermal imagery with LiDAR depth, though odometry synchronization is suboptimal, which may affect our odometry-based method. To better reflect the intended deployment scenario, we additionally evaluate on our custom dataset equipped with wheel odometry, which provides more reliable baseline estimates than GPS-based odometry that is susceptible to signal degradation and synchronization errors. The dataset comprises 2.8K urban roadside frames and 7K forest frames with RGB and thermal imagery.

\subsection{Evaluation Protocol}
Our proposed algorithm consists of two main components: (1)  a pose estimation module from optical flow with a metric baseline from odometry, and (2) a depth fusion stage that converts relative depth into metric depth. To validate pose accuracy, we evaluate triangulated depth quality against two baselines: a method that estimates pose using a metric depth model, and ground-truth pose as a reference. Unlike the former, our method requires only a metric baseline from any odometry source instead of a pretrained metric depth model. For depth estimation, we compare against both single-image and video-based metric depth models, including Video Depth Anything (VDA)~\cite{chen2025video}.
To assess temporal consistency, we additionally report the Temporal Alignment Error (TAE)~\cite{yang2024depthanyvideo}, defined as the bidirectional AbsRel between geometrically warped depth maps of consecutive frames:
$\mathrm{TAE}=\frac{1}{2(T-2)}\sum_{k=0}^{T-1}
\mathrm{AbsRel}\!\left(f(\hat{x}_{d}^{k},p^{k}),\hat{x}_{d}^{k+1}\right)
+\mathrm{AbsRel}\!\left(f(\hat{x}_{d}^{k+1},p_{-}^{k+1}),\hat{x}_{d}^{k}\right)$,
where $T$ is the number of frames, $f(\hat{x}_{d}^{k}, p^{k})$ denotes the depth map of frame $k$ projected into frame $k+1$ using relative pose $p^{k}$, and $p^{k+1}_{-}$ is the reverse projection. This metric reflects the consistency between two consecutive depth estimates.

\subsection{Triangulation Evaluation}

Accurate triangulation requires metric-scale pose estimation. MADPose~\cite{madpose} recovers metric-scale camera pose by jointly solving for relative pose and affine corrections to depth predictions from a pretrained metric depth model (UniDepth), making it dependent on the depth model's generalization capability. In contrast, our method uses a relative depth model only for depth structure, recovering metric scale solely from a single external odometry measurement—such as wheel odometry, VIO, or GPS—without depending on a metric depth model. We additionally include triangulation with ground-truth pose to illustrate the effect of temporal misalignment: in real sequences, imperfect synchronization between pose measurements and image frames can degrade triangulation accuracy, whereas in synthetic data such as TartanAir, perfect synchronization makes ground-truth pose the strongest baseline.

The results are summarized in Table~\ref{tab:tri}. On KITTI, where UniDepth generalizes well, MADPose achieves competitive triangulation accuracy comparable to ours. On out-of-distribution datasets—TartanAir, MS2, and our custom dataset—UniDepth performance degrades, and MADPose triangulation accuracy drops substantially as a result. Our method, relying only on optical flow and odometry, maintains consistently high accuracy across all datasets. On our custom mobile robot dataset equipped with wheel odometry, our method achieves approximately three times higher $\delta_1$ accuracy than MADPose.

\begin{table*}[t]
\centering
\resizebox{\textwidth}{!}{
\begin{tabular}{l ccc ccc ccc cc ccc ccc}
\toprule
& \multicolumn{11}{c}{\textbf{RGB}} & \multicolumn{6}{c}{\textbf{Thermal}} \\
\cmidrule(lr){2-12} \cmidrule(lr){13-18} 
& \multicolumn{3}{c}{KITTI} & \multicolumn{3}{c}{TartanAir} & \multicolumn{3}{c}{Roadside} & \multicolumn{2}{c}{Forest} & \multicolumn{3}{c}{MS2} & \multicolumn{3}{c}{Roadside} \\
\cmidrule(lr){2-4} \cmidrule(lr){5-7} \cmidrule(lr){8-10} \cmidrule(lr){11-12} \cmidrule(lr){13-15} \cmidrule(lr){16-18} 
Method
& AbsRel$\downarrow$ & $\delta\!<\!1.25\uparrow$ & TAE$\downarrow$
& AbsRel$\downarrow$ & $\delta\!<\!1.25\uparrow$ & TAE$\downarrow$
& AbsRel$\downarrow$ & $\delta\!<\!1.25\uparrow$ & TAE$\downarrow$
& AbsRel$\downarrow$ & $\delta\!<\!1.25\uparrow$
& AbsRel$\downarrow$ & $\delta\!<\!1.25\uparrow$ & TAE$\downarrow$
& AbsRel$\downarrow$ & $\delta\!<\!1.25\uparrow$ & TAE$\downarrow$
\\
\midrule
\midrule
\multicolumn{18}{l}{\textit{Full range (0--80\,m)}} \\
\midrule
UniDepth
& \best{0.047} & \best{0.977} & \best{4.34}
& \second{0.503} & 0.176 & 11.11
& \second{0.465} & \second{0.201} & 11.92
& \second{0.444} & 0.088
& \best{0.205} & \second{0.698} & 5.84
& \best{0.394} & \second{0.245} & 11.92
\\
DA v2 (metric)
& 0.171 & 0.773 & 5.21
& 0.513 & \second{0.372} & 5.53
& 0.494 & 0.177 & \second{3.28}
& \best{0.418} & \second{0.336}
& 0.405 & 0.187 & \second{4.87}
& \second{0.527} & 0.193 & \second{2.89}
\\
VDA
& 0.356 & 0.321 & \second{5.06}
& 0.599 & 0.342 & \best{4.44}
& 2.198 & 0.010 & \best{1.98}
& 1.339 & 0.041
& 0.590 & 0.078 & \best{3.67}
& 2.275 & 0.011 & \best{1.98}
\\
Ours
& \second{0.137} & \second{0.877} & 5.35
& \best{0.427} & \best{0.688} & \second{5.42}
& \best{0.309} & \best{0.725} & 5.27
& 0.480 & \best{0.520}
& \second{0.247} & \best{0.700} & 5.29
& 0.570 & \best{0.527} & 5.27
\\
\bottomrule
\end{tabular}}
\caption{\textbf{Depth estimation and temporal consistency.} 
\best{Bold}: best, \second{underline}: second best per dataset. VDA~\cite{chen2025video} is a video-based depth model evaluated for temporal consistency (TAE); DA~v2 metric uses the publicly available outdoor model fine-tuned on Virtual KITTI.}\label{tab:depthest}
\end{table*}
\subsection{Depth Estimation Evaluation}
We compare our method against single-image metric depth models and a video-based depth model across all five datasets. Since all evaluated datasets cover outdoor scenes, we use the publicly available outdoor variant of DA v2 metric, fine-tuned on Virtual KITTI. Results are summarized in Table~\ref{tab:depthest}.

UniDepth achieves the highest accuracy on KITTI, where its training distribution closely matches the evaluation domain. Our method, which uses a relative depth model not trained on KITTI, nonetheless achieves competitive performance on this benchmark. A key advantage of using a relative depth model is that sky regions are naturally suppressed and object boundaries are well-preserved, as shown in Fig.~\ref{fig:rgbt}. On MS2 and our custom dataset—which include challenging conditions such as thermal imagery and dense forest scenes—our method achieves the highest overall accuracy, demonstrating strong generalization to out-of-distribution environments. VDA achieves favorable temporal consistency (TAE) on in-distribution datasets, but its metric accuracy degrades significantly in out-of-distribution sequences, as video depth models are not designed for metric scale recovery. Note that a systematically biased depth estimate—such as a constant underestimation—can yield artificially low TAE, since temporal consistency measures frame-to-frame coherence rather than absolute accuracy; a consistently wrong prediction is still temporally consistent. Our method maintains competitive TAE while providing reliable metric depth across all evaluated environments.

The temporal consistency of our method is further illustrated in Figs.~\ref{fig:pt}--\ref{fig:absrel}. The accumulated 3D point cloud in Fig.~\ref{fig:pt} demonstrates that temporally consistent depth predictions yield high-precision reconstruction when multiple frames are integrated. UniDepth, lacking temporal consistency, produces accumulated point clouds where object shapes are difficult to discern, whereas our method clearly reconstructs lane structures and building geometry, as shown in Fig.~\ref{fig:acc}. Fig.~\ref{fig:absrel} shows per-frame absolute error on our custom urban roadside dataset, where our method maintains consistently low error compared to the substantial frame-to-frame deviations exhibited by UniDepth.

We further analyze depth accuracy across distance ranges in Table~\ref{tab:nearfar}. In the near range (0--20\,m), our method achieves the highest accuracy on all out-of-distribution datasets, benefiting from well-conditioned triangulation geometry. In the far range (20--80\,m), triangulation becomes less effective due to diminishing parallax, and single-image metric models such as DA~v2 perform comparably or better. This is a fundamental limitation of geometry-based scaling shared by all monocular triangulation methods.

\begin{table}[t]
\centering
\resizebox{\columnwidth}{!}{
\begin{tabular}{l cc cc cc cc}
\toprule
& \multicolumn{2}{c}{\textbf{KITTI}}
  & \multicolumn{2}{c}{\textbf{TartanAir}}
  & \multicolumn{2}{c}{\textbf{Roadside}}
  & \multicolumn{2}{c}{\textbf{Forest}} \\
\cmidrule(lr){2-3} \cmidrule(lr){4-5} \cmidrule(lr){6-7} \cmidrule(lr){8-9} 
Method
  & AbsRel$\downarrow$ & $\delta\!<\!1.25\uparrow$
  & AbsRel$\downarrow$ & $\delta\!<\!1.25\uparrow$
  & AbsRel$\downarrow$ & $\delta\!<\!1.25\uparrow$
  & AbsRel$\downarrow$ & $\delta\!<\!1.25\uparrow$
\\
\midrule
\multicolumn{9}{l}{\textit{Near range (0--20\,m)}} \\
\midrule
UniDepth
  & \best{0.039} & \best{0.989}
  & \second{0.485} & 0.202
  & \second{0.432} & \second{0.241}
  & \second{0.428} & 0.096 \\
DA v2 (metric)
  & 0.154 & 0.821
  & 0.647 & 0.322
  & 0.508 & 0.168
  & 0.435 & \second{0.331} \\
VDA
  & 0.313 & 0.376
  & 0.575 & \second{0.341}
  & 1.369 & 0.106
  & 1.081 & 0.105 \\
Ours
  & \second{0.128} & \second{0.880}
  & \best{0.339} & \best{0.712}
  & \best{0.165} & \best{0.860}
  & \best{0.304} & \best{0.617} \\
\midrule
\multicolumn{9}{l}{\textit{Far range (20--80\,m)}} \\
\midrule
UniDepth
  & \best{0.075} & \best{0.940}
  & 0.430 & 0.232
  & \second{0.374} & 0.323
  & \second{0.480} & 0.084 \\
DA v2 (metric)
  & \second{0.182} & 0.732
  & \best{0.271} & \second{0.447}
  & \best{0.244} & \best{0.509}
  & \best{0.214} & \best{0.558} \\
VDA
  & 0.408 & 0.339
  & 0.448 & 0.367
  & 1.488 & 0.047
  & 0.881 & 0.124 \\
Ours
  & 0.205 & \second{0.749}
  & \second{0.415} & \best{0.559}
  & 0.570 & \second{0.361}
  & 0.780 & \second{0.228} \\
\bottomrule
\end{tabular}}
\caption{\textbf{Depth estimation accuracy across near and far ranges.} We separately evaluate near (0–20 m) and far (20–80 m) regions on RGB sequences.}\label{tab:nearfar}
\end{table}

\begin{table}[t]
\centering
\resizebox{\columnwidth}{!}{
\begin{tabular}{l ccc ccc ccc}
\toprule
& \multicolumn{3}{c}{KITTI} & \multicolumn{3}{c}{Roadside} & \multicolumn{3}{c}{TartanAir} \\
\cmidrule(lr){2-4} \cmidrule(lr){5-7} \cmidrule(lr){8-10}
Method & AbsRel & $\delta\!<\!1.25$ & TAE & AbsRel & $\delta\!<\!1.25$ & TAE & AbsRel & $\delta\!<\!1.25$ & TAE \\
\midrule
w/o fusion   & 0.088 & 0.945 & 6.28 & 0.428 & 0.662 & 12.99 & - & 0.517 & 7.84 \\
w/o segment  & 0.088 & 0.952 & \best{4.38} & \second{0.270} & \best{0.746} & \second{6.16} & \second{0.274} & \second{0.685} & 5.40 \\
\textbf{Full} & \best{0.087} & \best{0.953} & \second{4.79} & \best{0.265} & \second{0.743} & 6.27 & \best{0.218} & \best{0.714} & \best{4.65} \\
\bottomrule
\end{tabular}
}
\caption{\textbf{Ablation study.} Effect of removing Bayesian temporal fusion (w/o fusion) and segment-wise scale consolidation (w/o segment). Each dataset is evaluated on a single representative sequence.}\label{tab:ablation}
\end{table}

\subsection{Ablation Study}
We ablate two key components on KITTI, TartanAir, and Roadside (RGB): (1) removing the Bayesian temporal fusion (w/o fusion), and (2) removing the segment-wise scale consolidation (w/o segment). As shown in Table~\ref{tab:ablation}, the fusion module plays a critical role in reducing TAE. Note that AbsRel is not reported for w/o fusion on TartanAir, as metric scale cannot be recovered without the fusion stage. The segment-wise consolidation further improves accuracy, particularly in geometrically diverse environments such as TartanAir.

\subsection{Failure Cases}
Our method can fail under two conditions. First, when camera motion is dominated by forward translation with minimal lateral displacement, as in high-speed highway driving, triangulation becomes degenerate and reliable depth observations cannot be obtained. Second, when dynamic objects occupy a large portion of the image (over 50\%), optical flow estimation becomes unreliable, degrading pose estimation and subsequent depth fusion.

\section{Conclusion}

In this paper, we introduce an algorithm that leverages wheel odometry to achieve metric depth estimation with strong temporal consistency, generalizing to unseen environments without additional training. By integrating relative depth from a foundation model trained on large-scale data with sparse depth obtained from optical flow and wheel odometry, our method produces accurate and reliable metric depth estimates. We evaluated our approach on KITTI, TartanAir, and MS2 datasets, and additionally collected approximately 10k frames in urban and forest environments to validate zero-shot generalization. The results demonstrate stable performance even in completely unseen environments.

For future work, we aim to address performance degradation in rotation-dominant segments, where epipolar geometry degenerates, and to investigate learned geometric reasoning that improves robustness in such cases while preserving zero-shot generalizability.

\paragraph{Acknowledgements.}
This work was supported in part by the Unmanned Vehicles Core Technology Research and Development Program through the National Research
Foundation of Korea (NRF) and Unmanned Vehicle Advanced Research Center (NRF2020M3C1C1A01086411); and in part by the National Research Foundation of Korea (NRF) grant (No.RS2023-00213897);  in part by Institute of Information \& communications Technology Planning \& Evaluation (IITP) under the Artificial Intelligence Convergence Innovation Human Resources Development (IITP-2026-RS-2023-00255968) grant funded by the Korea government(MSIT)
{
    \small
    \bibliographystyle{ieeenat_fullname}
    \bibliography{main}
}

\clearpage
\setcounter{section}{0}
\setcounter{figure}{0}
\renewcommand{\thefigure}{S\arabic{figure}}
\setcounter{table}{0}
\renewcommand{\thetable}{S\arabic{table}}

\twocolumn[{%
\renewcommand\twocolumn[1][]{#1}%
\newpage
\null
\vskip .375in
\begin{center}
  {\Large \bf PTC-Depth: Pose-Refined Monocular Depth Estimation\\
  with Temporal Consistency\\
  {\large Supplementary Material} \par}
  \vspace*{24pt}
  {
    \large
    \lineskip .5em
    \begin{tabular}[t]{c}
    \end{tabular}
    \par
  }
  \vskip .5em
  \vspace*{12pt}
\end{center}
}]

\section{Dataset and Sensor Specifications}
\label{sec:supp_dataset}

Our custom dataset is collected using multiple sensors mounted on a
\textit{Hunter SE} mobile platform, as shown in
Fig.~\ref{fig:supp_platform}. The RGB camera is an \textit{Intel
RealSense D455}, capturing images at a resolution of $640\times480$ at
60\,fps. The thermal camera is a \textit{FLIR Boson 640}, providing
LWIR data at a resolution of $640\times512$ at 60\,fps. Depth ground
truth is obtained using a \textit{Livox HAP} LiDAR. In addition, wheel
odometry from the platform is used to estimate the vehicle motion. The specifications of each sensor are summarized in
Fig.~\ref{fig:supp_platform}.

\newsavebox{\sensorbox}
\savebox{\sensorbox}{%
\begin{minipage}{0.55\columnwidth}
    \centering
    \includegraphics[width=\textwidth]{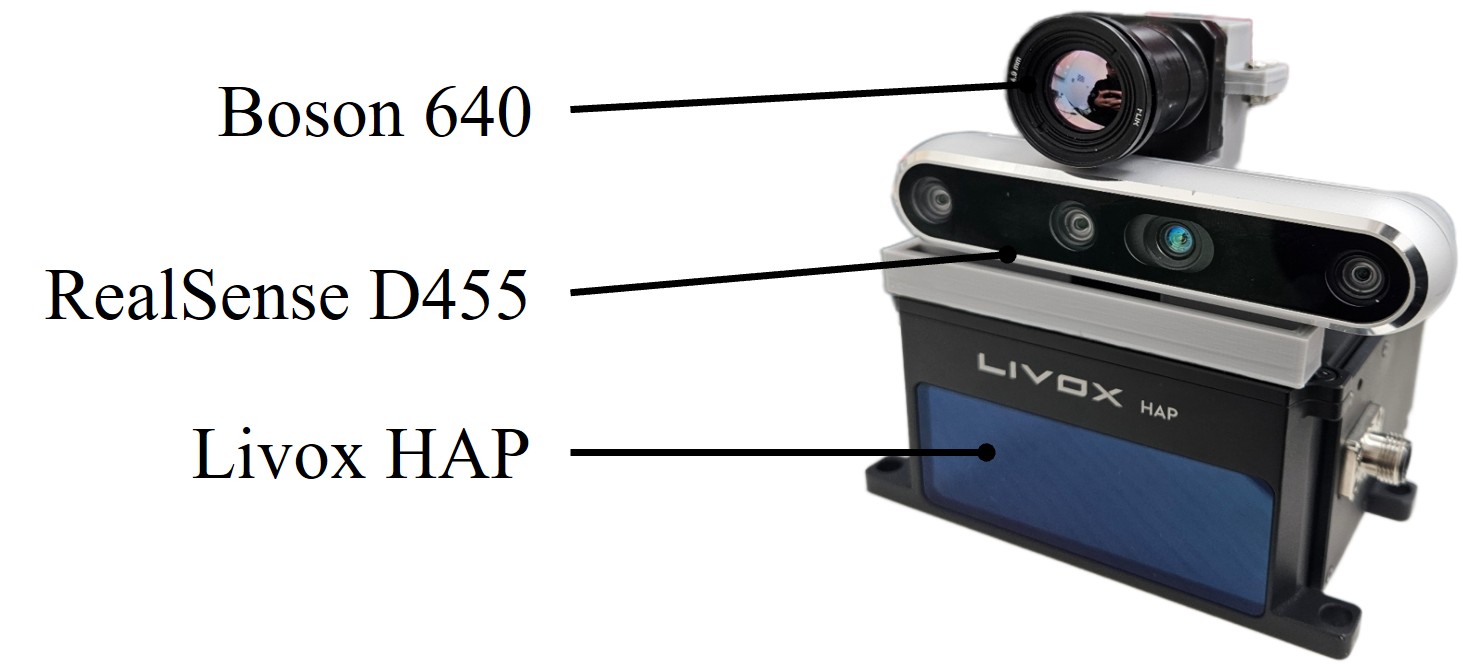}\\[-0.2em]
    \resizebox{\textwidth}{!}{
    \begin{tabular}{l|cc}
    \hline
    \textbf{\begin{tabular}[c]{@{}l@{}}Sensor type\\ (Model)\end{tabular}} & \textbf{Frame rate} & \textbf{Specification}\\ \hline
    \begin{tabular}[c]{@{}l@{}}RGB Camera\\ (Intel RealSense D455)\end{tabular} & 60 fps & \begin{tabular}[c]{@{}c@{}}640$\times$480\\ FOV: 90$^\circ$(H)$\times$65$^\circ$(V)\\ Global Shutter\end{tabular}\\ \hline
    \begin{tabular}[c]{@{}l@{}}Thermal Camera\\ (FLIR Boson 640)\end{tabular} & 60 fps & \begin{tabular}[c]{@{}c@{}}640$\times$512\\ FOV: 95$^\circ$(H)$\times$82$^\circ$(V)\\ LWIR (8--14 $\mu$m)\end{tabular}\\ \hline
    \begin{tabular}[c]{@{}l@{}}LiDAR\\ (Livox HAP)\end{tabular} & 10 fps & \begin{tabular}[c]{@{}c@{}}Range: 150m\\ FOV: 120$^\circ$(H)$\times$25$^\circ$(V)\end{tabular}\\ \hline
    \end{tabular}}
\end{minipage}}

\begin{figure}[H]
\centering
\usebox{\sensorbox}%
\hfill
\begin{minipage}[c]{0.42\columnwidth}
    \centering
    \includegraphics[height=\dimexpr\ht\sensorbox+\dp\sensorbox\relax,width=\textwidth,keepaspectratio]{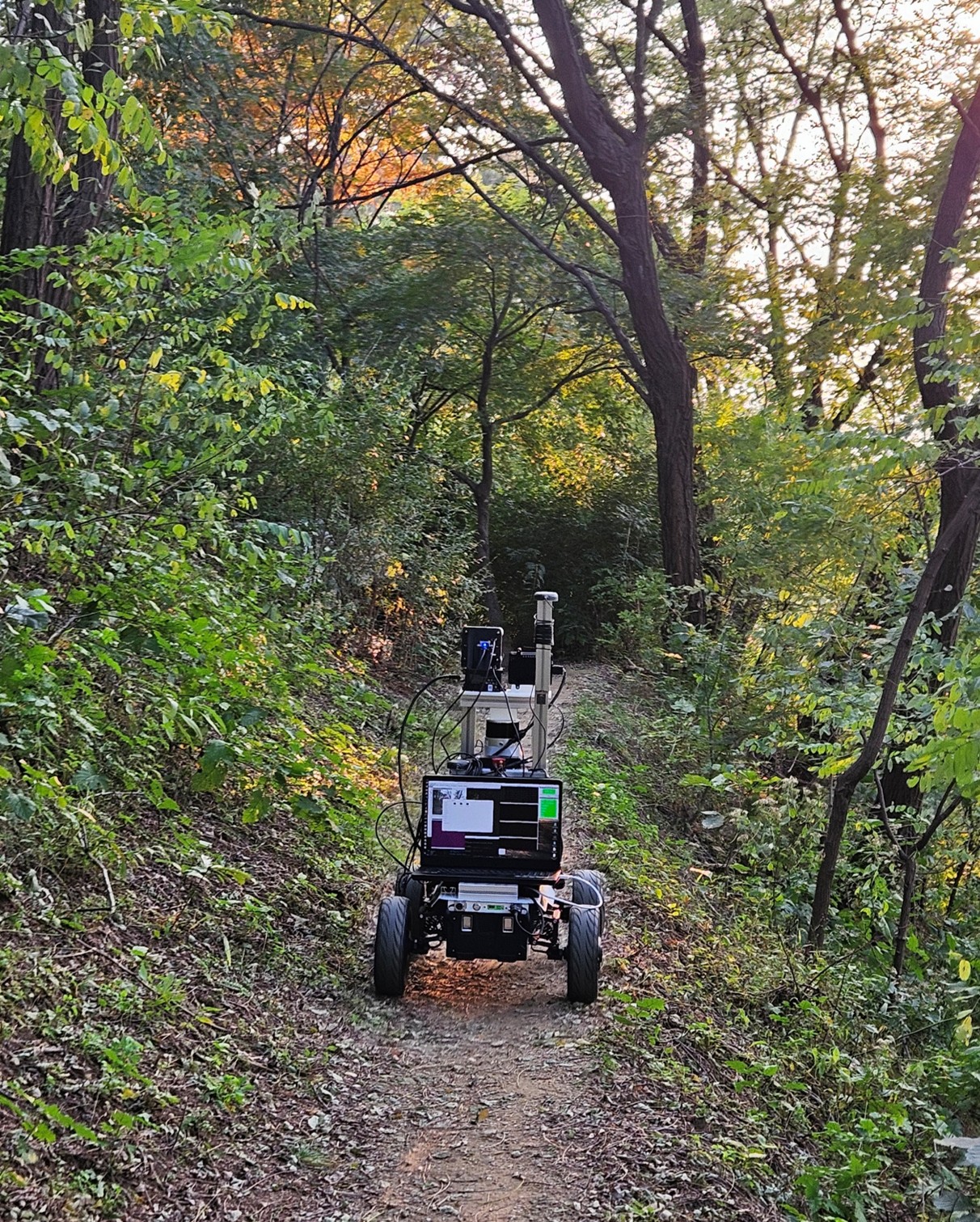}
\end{minipage}
\caption{\textbf{Data collection platform.}
(Left)~Sensor module with FLIR Boson~640, Intel RealSense D455,
and Livox HAP LiDAR.
(Right)~Hunter SE platform in a forest environment.}
\label{fig:supp_platform}
\end{figure}

\section{Overview}
\begin{figure*}[t]
    \centering
    \includegraphics[width=\textwidth]{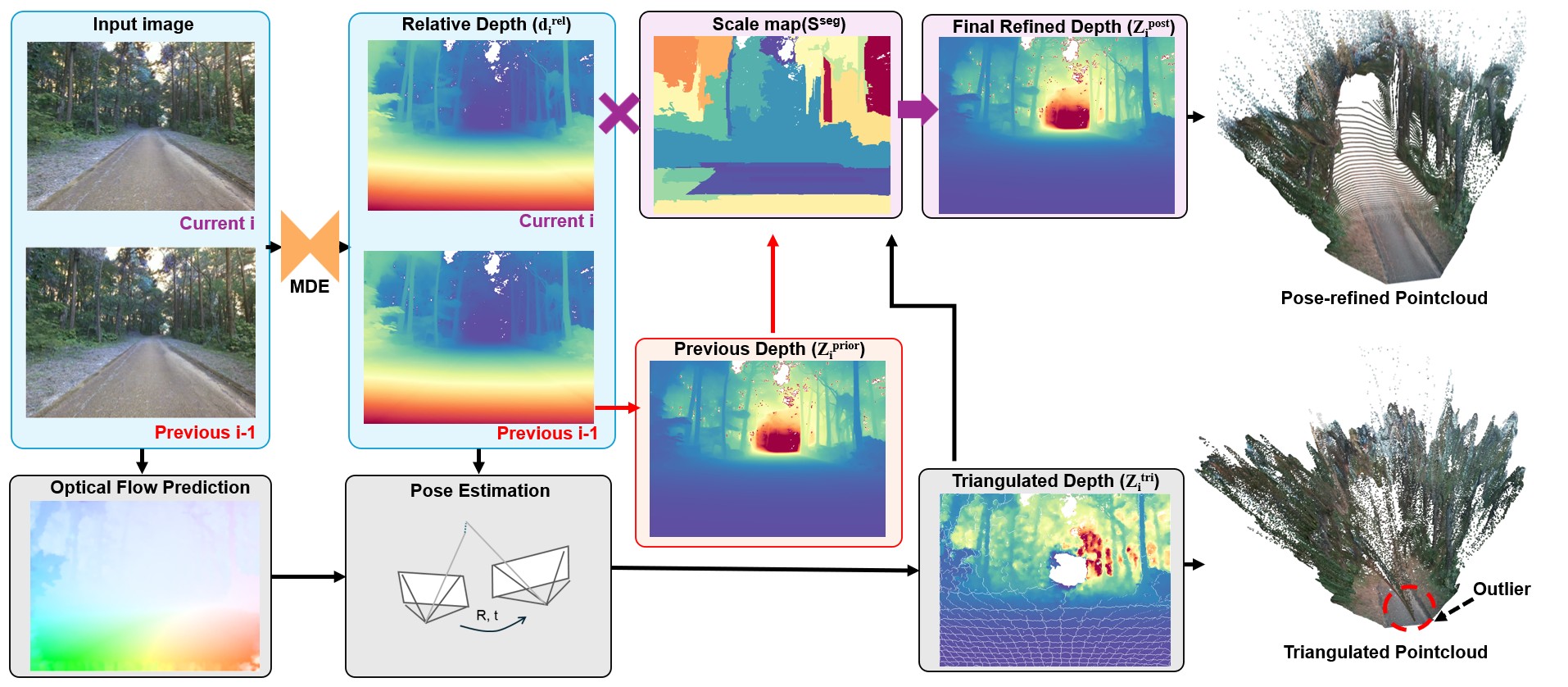}
    \caption{\textbf{Overall Process of Our Proposed Algorithm}}
    \label{fig:pip}
\end{figure*}

This supplementary document provides implementation details and additional analyses not included in the main paper. We first describe our custom dataset and sensor specifications (Section~\ref{sec:supp_dataset}), followed by the procedures used for robust motion estimation (Section~\ref{sec:supp_motion}), Sampson residual computation and Bayesian variance modeling (Section~\ref{sec:supp_triangulation}), superpixel-wise scale refinement (Section~\ref{sec:supp_superpixel}), an ablation on odometry accuracy (Section~\ref{sec:supp_ablation}), and runtime analysis (Section~\ref{sec:supp_runtime}). Additional limitations are discussed in Section~\ref{sec:supp_limitations}. The overall process is shown in Fig.~\ref{fig:pip}.

\section{Robust Motion Estimation}
\label{sec:supp_motion}

This section provides additional implementation details for the motion-estimation stage.
The main paper introduces the motion-field formulation; here, we describe how correspondences
are selected, how flow--motion consistency is evaluated, and how rotation-dominant frames
are handled in practice.

\subsection{Normalized motion-field residual}

Let $\mathbf{f}(x)$ denote the observed optical flow at pixel $x$ and
$\dot{\mathbf{p}}(x)$ the motion-field prediction for parameters
$(\boldsymbol{\Omega},\boldsymbol{T})$.
We first compute the pixel-domain discrepancy
\begin{equation}
r(x)=\left\|\mathbf{f}(x)-\dot{\mathbf{p}}(x)\right\|_2 .
\end{equation}
Because the magnitude of $\mathbf{f}(x)$ varies across the image, we normalize this residual as
\begin{equation}
e(x)=
\frac{
    r(x)
}{
    \max\!\left(\|\mathbf{f}(x)\|_2,\;\tau\right)
}.
\end{equation}
Because $\|\mathbf{f}(x)\|_2$ can be very small in near-static regions,
we set $\tau$ as a small lower bound (e.g., $\tau = 1$ pixel) to ensure numerical stability.
This relative residual provides a scale-invariant measure of consistency and yields a
stable inlier criterion across both large-motion and near-static regions.

\subsection{Spatially and depth-balanced sampling}

To avoid bias toward locally dense textures, we draw correspondences from a stratified distribution:
(i) the image is divided into coarse spatial regions;
(ii) candidate pixels are grouped into a few depth intervals; and
(iii) each spatial--depth group contributes at most a fixed number of samples.
All selected pixels must have finite flow, finite inverse depth, and non-negligible flow magnitude.
This produces a well-conditioned set of correspondences supporting motion estimation across the full field of view.

\subsection{Adaptive RANSAC with directional consistency}

RANSAC evaluates hypotheses using the normalized residual $e(x)$.
Two additional robustness mechanisms are employed.

\paragraph{Directional consistency.}
For sufficiently large flows, we measure the angular deviation
\begin{equation}
\Delta\theta(x)=
\arccos\!\left(
\frac{
\mathbf{f}(x)\cdot\dot{\mathbf{p}}(x)
}{
\|\mathbf{f}(x)\|_2\,\|\dot{\mathbf{p}}(x)\|_2
}
\right).
\end{equation}
Correspondences with abnormally large deviations are rejected using a robust
threshold derived from the distribution of $\Delta\theta(x)$.
This removes flow outliers whose direction is inconsistent with rigid motion.

\paragraph{Adaptive residual threshold.}
The inlier threshold $\eta$ for $e(x)$ is initialized from robust
statistics of the residual distribution:
\begin{equation}
\eta_{0} = \mathrm{median}(e) + \lambda\;\mathrm{MAD}(e),
\end{equation}
where $\mathrm{MAD}(e) = \mathrm{median}(|e - \mathrm{median}(e)|)$
and $\lambda$ controls the tightness of the criterion.
After each hypothesis, $\eta$ is adjusted: if the inlier ratio falls
below a target, $\eta$ is relaxed; otherwise it is tightened.
Hypotheses are scored by both inlier count and spatial coverage, favoring
solutions supported across the image.

\subsection{IRLS refinement}

After RANSAC selects the best hypothesis, we refine the parameters
$(\boldsymbol{\Omega},\boldsymbol{T})$ using a small number of
iteratively re-weighted least squares (IRLS) iterations.
Each inlier pixel $x$ is assigned a Huber weight
\begin{equation}
w(x) =
\begin{cases}
1, & e(x) \leq \eta, \\[2pt]
\eta\,/\,e(x), & e(x) > \eta,
\end{cases}
\end{equation}
where $\eta$ is the inlier threshold from the RANSAC stage.
At each iteration, the weighted linear system derived from the
motion-field equation (Eq.~1 of the main paper) is solved to update
$(\boldsymbol{\Omega},\boldsymbol{T})$, and the weights are
recomputed.  This refinement stabilizes the estimated rotation and
translation and suppresses residual outliers that survive RANSAC
sampling.

\subsection{Inlier validation and flow fusion}

Given the final estimate $(\boldsymbol{\Omega}^*,\boldsymbol{T}^*)$, each pixel is
validated using both $e(x)$ and $\Delta\theta(x)$.
Pixels satisfying both criteria retain their observed flow; others are replaced
with the motion-field prediction.
This fused flow prevents a small set of erroneous flows from influencing
triangulation or scale estimation.

\section{Triangulation and Bayesian Fusion Details}
\label{sec:supp_triangulation}

This section provides the closed-form expressions for the Sampson
residual map and the consistency score that are referenced in the main
paper but omitted for brevity.

\subsection{Sampson Residual Map}
\label{sec:supp_sampson}

Given the estimated relative pose
$(\boldsymbol{\Omega},\boldsymbol{T})$, we construct the fundamental
matrix
\begin{equation}
F = K^{-\top}[\boldsymbol{T}]_\times \boldsymbol{\Omega}\, K^{-1},
\end{equation}
where $[\boldsymbol{T}]_\times$ denotes the skew-symmetric matrix of
$\boldsymbol{T}$ and $K$ is the camera intrinsic matrix.

For each correspondence
$(\mathbf{x}_{i-1},\,\mathbf{x}_{i})$ between frames $i{-}1$ and $i$,
the Sampson residual is computed as
\begin{equation}
\rho(x) =
\frac{
  \left(\mathbf{x}_{i}^{\top}\,F\,\mathbf{x}_{i-1}\right)^{2}
}{
  (F\,\mathbf{x}_{i-1})_1^2 + (F\,\mathbf{x}_{i-1})_2^2
  + (F^{\top}\mathbf{x}_{i})_1^2 + (F^{\top}\mathbf{x}_{i})_2^2
},
\label{eq:sampson}
\end{equation}
where $(\cdot)_j$ selects the $j$-th component. Each $\rho(x)$ is
assigned to its corresponding pixel in frame~$i$, forming the
\emph{Sampson residual map}. This map is used both to derive the
per-pixel observation variance $V^{\mathrm{obs}}$ and to inflate the
prior variance $V^{\mathrm{prior}}$, as described in the main paper.

\subsection{Consistency Score}
\label{sec:supp_consistency}

To prevent aggressive Kalman updates where triangulation and prior
disagree, we compute a per-pixel consistency score $c(x)\in[0,1]$.
The relative discrepancy between the observed and prior scales is
\begin{equation}
\delta(x) = \frac{|S^{\mathrm{obs}}(x)-S^{\mathrm{prior}}(x)|}
                  {S^{\mathrm{obs}}(x)},
\end{equation}
and the frame-level tolerance is estimated as
$\sigma_e = \mathrm{MAD}(\delta)$, smoothed across time with an
exponential moving average. The consistency score is then
\begin{equation}
c(x) = \exp\!\left(-\frac{\delta(x)^{2}}{2\,\sigma_e^{2}}\right).
\end{equation}
Pixels with $c(x)\approx 1$ receive the full Kalman gain, while pixels
with low $c(x)$ are down-weighted to suppress unreliable triangulation.
This score is used to cap the raw Kalman gain $\kappa_{\mathrm{raw}}$
as described in the main paper.

\section{Superpixel-Based Spatial Refinement}
\label{sec:supp_superpixel}

Although the Bayesian update described in the main paper provides a
pixel-wise fusion of the triangulated depth and the warped prior,
residual spatial inconsistencies remain due to imperfect optical flow,
unstable triangulation, and amplified noise in distant regions.
To mitigate these effects, we refine the posterior scale field
$S^{\mathrm{post}}$ using superpixels that follow both the color and
geometric structure of the scene.

\subsection{Motivation}

Neither triangulation nor prior warping is perfectly reliable.
Triangulated depths become unstable when the stereo baseline is small,
the motion is rotation-dominant, or when the flow correspondence lies
near occlusion boundaries.
Similarly, the optical flow used for warping may contain several-pixel
errors in low-texture regions or around dynamic objects. These errors
propagate into the posterior $S^{\mathrm{post}}(x)$ and appear as
irregular islands or thin artifacts after fusion.

Pixel-wise Bayesian fusion suppresses high-frequency noise but cannot
fully remove spatially coherent distortions caused by flow mismatch or
triangulation failures. A structural, surface-level correction stage is
therefore necessary.
\begin{figure}[t]
    \centering
    \includegraphics[width=\columnwidth]{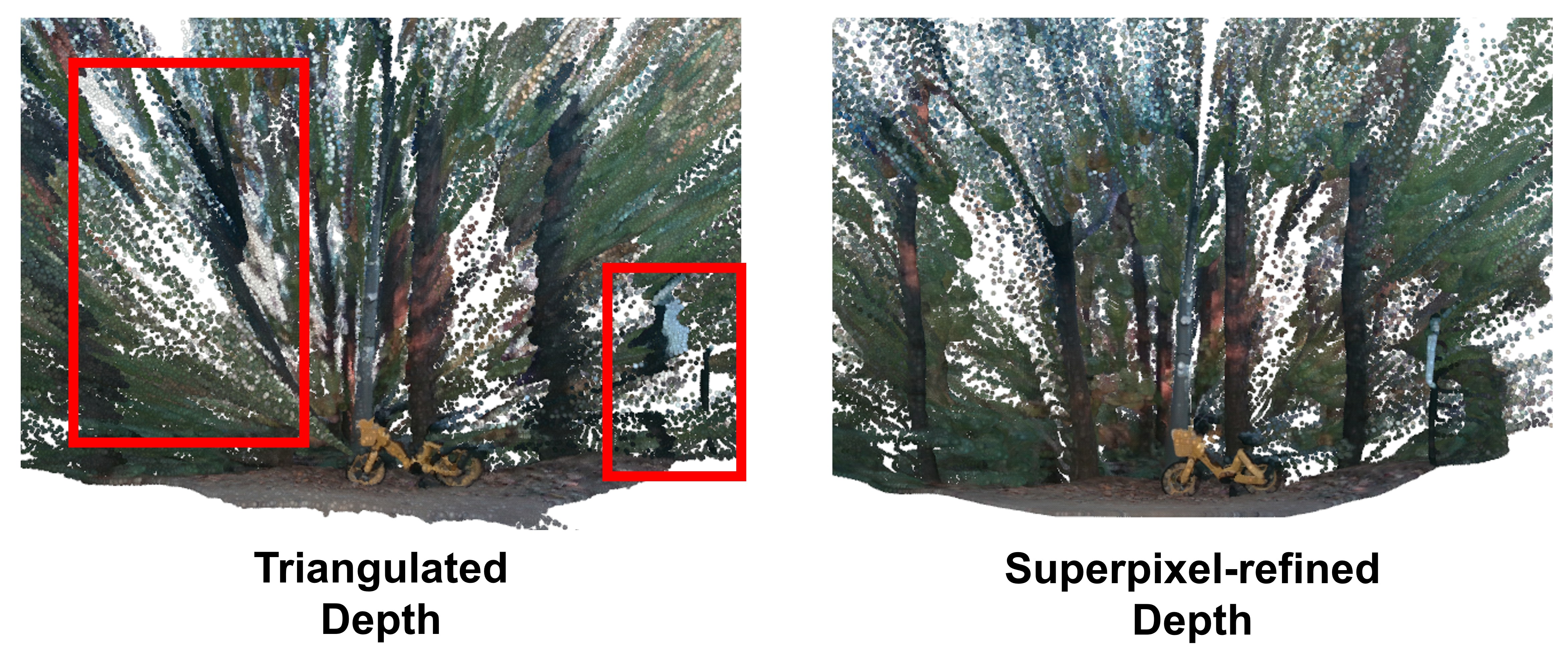}
    \caption{\textbf{Comparison of triangulated vs.~refined point clouds.}
    (Left) Raw triangulated depth produces scattered points and surface
    discontinuities due to imperfect flow and unstable geometry.
    (Right) After our Bayesian update and superpixel refinement, the
    reconstructed surfaces become smoother and more consistent.}
    \label{fig:supp_pc_compare}
\end{figure}

Figure~\ref{fig:supp_pc_compare} illustrates this issue: the raw
triangulated point cloud (left) contains numerous spikes and scattered
surface fragments, while our refined point cloud (right) is
significantly smoother and geometrically more coherent.

\subsection{LAB-based segmentation aligned to depth structure}

We apply Felzenszwalb segmentation~\cite{felzenszwalb2004efficient}
using LAB color and relative depth features as input, rather than RGB
alone, to better align superpixel boundaries with geometric structure. The LAB representation is chosen because the $L$ channel is less
sensitive to illumination changes, while the $a$ and $b$ channels provide
stable chromatic cues. Combined with relative depth edges, the resulting
superpixels adhere closely to object boundaries and planar surfaces.

\begin{figure}[t]
    \centering
    \includegraphics[width=\columnwidth]{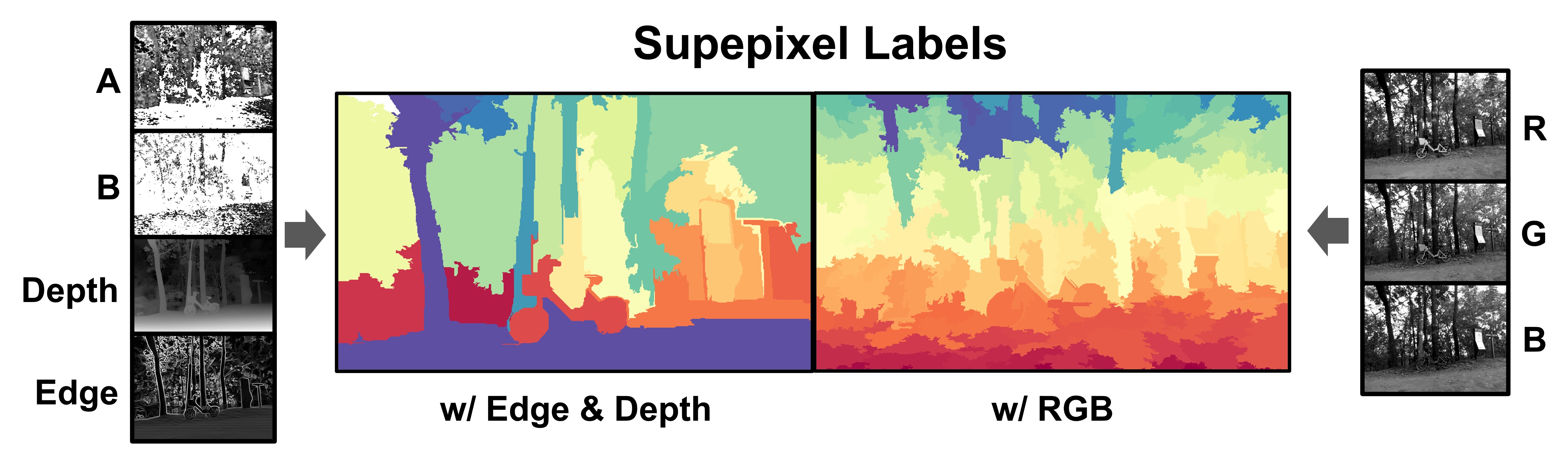}
    \caption{\textbf{LAB-based vs.~RGB superpixel segmentation.}
    (Left) Our LAB+depth segmentation follows true object and geometric
    contours, making each label more consistent with the underlying
    3D structure.
    (Right) RGB-based segmentation often misaligns with physical surface.}
    \label{fig:supp_seg_compare}
\end{figure}

Figure~\ref{fig:supp_seg_compare} compares segmentation results obtained
from the original RGB image and from our LAB+depth representation.
While RGB-based superpixels tend to bleed across surfaces, the LAB-based
segmentation produces spatially coherent, semantically meaningful
regions that align with depth discontinuities.

\subsection{Label-wise scale refinement}

Let $\{\Lambda_\ell\}$ denote the set of superpixel segments.
For each segment $\Lambda_\ell$, we refine the pixel-wise posterior scale by

\[
\bar{s}_\ell
= \mathrm{median}\,\{\,S^{\mathrm{post}}_k \mid k \in \Lambda_\ell \,\}.
\]

This median-based representative scale is then applied to all pixels in
the corresponding superpixel:

\[
S^{\mathrm{seg}}_k =
\begin{cases}
\bar{s}_\ell, & k \in \Lambda_\ell, \\
S^{\mathrm{post}}_k, & \text{otherwise}.
\end{cases}
\]

Because all pixels in $\Lambda_\ell$ originate from the same physical surface,
the above operation enforces geometric consistency across the region
while automatically rejecting local artifacts caused by noisy flow or
triangulation.

This refinement offers the following advantages:

\begin{itemize}
    \item \textbf{Flow robustness:} A few-pixel flow mismatch no longer
    produces ripples or tearing across a surface because the label
    enforces a single representative scale.
    \item \textbf{Triangulation stability:} Noisy triangulated points
    within a label are suppressed when they are inconsistent with the
    dominant depth trend of the surface.
    \item \textbf{Structure preservation:} Since superpixels align with
    depth boundaries (Fig.~\ref{fig:supp_seg_compare}), refinement does
    not blur across object edges.
    \item \textbf{Noise reduction in distant regions:}
    long-range triangulation noise is absorbed into the label median,
    producing smoother geometry as seen in
    Fig.~\ref{fig:supp_pc_compare}.
\end{itemize}

The refined scale field $S^{\mathrm{seg}}$ is subsequently used to compute
the final metric depth:
\[
z^{\mathrm{post}}_k
= S^{\mathrm{seg}}_k \cdot d^{\mathrm{rel}}_k.
\]

This stage acts as a structural regularizer that complements the
Bayesian update and stabilizes the depth estimates across large spatial
regions.

\section{Ablation on Odometry Accuracy}
\label{sec:supp_ablation}

To investigate the sensitivity of our method to odometry quality, we
replace the estimated pose with ground-truth (GT) components and measure
the effect on depth accuracy and temporal consistency. Three settings
are compared: (1)~fully estimated rotation and translation from optical
flow, (2)~GT rotation with estimated translation, and (3)~full GT
pose. Results are summarized in Table~\ref{tab:gt_ablation}.

\begin{table}[h]
\centering
\resizebox{\columnwidth}{!}{
\begin{tabular}{l ccc ccc}
\toprule
& \multicolumn{3}{c}{TartanAir (Synthetic RGB)}
& \multicolumn{3}{c}{MS2 (Thermal)} \\
\cmidrule(lr){2-4} \cmidrule(lr){5-7}
Pose setting
& AbsRel$\downarrow$ & $\delta\!<\!1.25\uparrow$ & TAE$\downarrow$
& AbsRel$\downarrow$ & $\delta\!<\!1.25\uparrow$ & TAE$\downarrow$ \\
\midrule
Estimated $\boldsymbol{\Omega}$, $\boldsymbol{T}$
& 0.218 & 0.714 & 4.65
& 0.513 & 0.603 & 5.75 \\
GT $\boldsymbol{\Omega}$ only
& 0.183 & 0.755 & 4.45
& 0.254 & 0.624 & 5.91 \\
GT $\boldsymbol{\Omega}$, $\boldsymbol{T}$
& 0.172 & 0.799 & 4.19
& 0.974 & 0.370 & -- \\
\bottomrule
\end{tabular}}
\caption{\textbf{Ablation on odometry accuracy.}
TartanAir: neighborhood (4.2K frames),
MS2: 2021-08-13-21-18-04 (10K frames).
On TartanAir, GT pose consistently improves accuracy thanks to perfect
synchronization. On MS2, full GT pose degrades performance due to
temporal misalignment between thermal images and RTK-GPS.}
\label{tab:gt_ablation}
\end{table}

On TartanAir, where image--pose synchronization is perfect, using GT
pose consistently improves both accuracy and temporal consistency.
However, on MS2, using full GT pose actually degrades performance. This
is because the thermal images and RTK-GPS measurements are not
perfectly synchronized, and the resulting temporal misalignment
introduces errors in triangulation that outweigh the benefit of more
accurate pose. This result highlights that our flow-based pose
estimation is more robust to synchronization issues than directly using
external pose measurements.

\section{Runtime Analysis}
\label{sec:supp_runtime}

We profile our method on a desktop PC equipped with an AMD Ryzen~9
9900X CPU. Table~\ref{tab:runtime} summarizes the per-frame
computational breakdown at KITTI resolution ($376 \times 1241$).

\begin{table}[h]
\centering
\resizebox{\columnwidth}{!}{
\begin{tabular}{lccccc|c}
\toprule
 & Seg+Flow & Motion & Scale & Tri+Fusion & \textbf{C++ Total} & \textbf{+Depth} \\
\midrule
Time (ms) & 43 & 25 & 15 & 6 & \textbf{90} & \textbf{156} \\
\bottomrule
\end{tabular}}
\caption{\textbf{Runtime breakdown at KITTI resolution
($376 \times 1241$).} The full pipeline, including monocular depth
estimation, achieves \textbf{6.4\,FPS}.}
\label{tab:runtime}
\end{table}

The C++ core (segmentation, optical flow, motion estimation, scale
recovery, triangulation, and Bayesian fusion) runs in approximately
90\,ms per frame. Including the monocular relative depth model
(DepthAnything~v2), the total processing time is 156\,ms per frame
(6.4\,FPS), which is suitable for real-time robotics applications.

In terms of memory, the pipeline maintains approximately 100\,MB of
active and persistent buffers per megapixel of input resolution. The
dominant allocations are per-pixel float32 maps for depth, scale,
variance, and optical flow, along with persistent state buffers that
carry the posterior estimate across frames.

\section{Limitations}
\label{sec:supp_limitations}
\begin{figure*}
  \centering
  \includegraphics[width=0.8\textwidth]{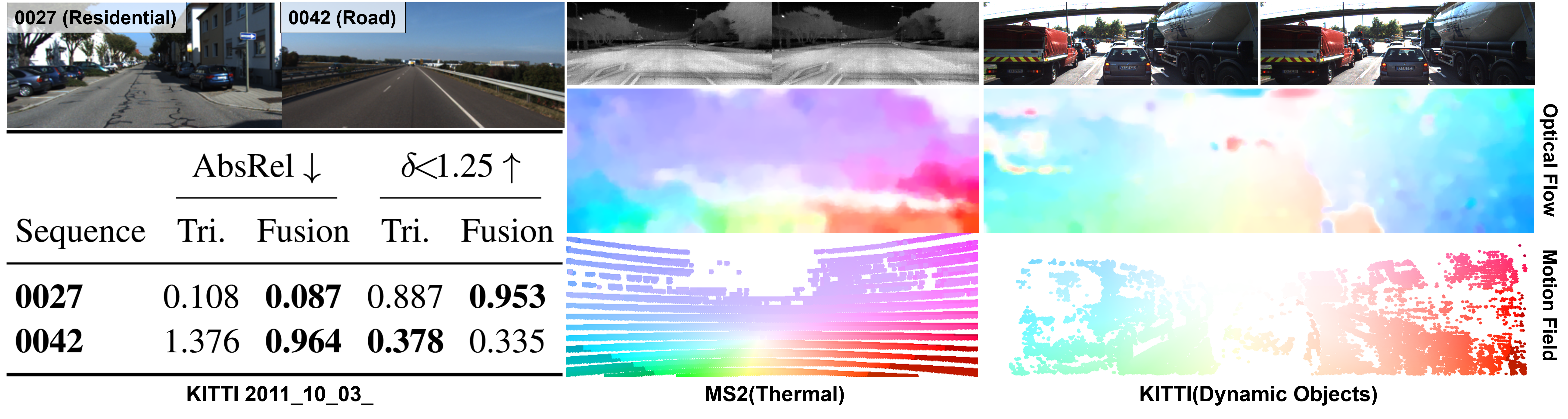}
  \caption{
    \textbf{Failure cases.}
    (Left) Forward-dominant motion in KITTI Seq~42 causes degenerate
    triangulation, resulting in unreliable depth observations.
    (Right) Scenes with dominant dynamic objects produce inconsistent
    optical flow, which degrades pose estimation and subsequent depth
    fusion.
  }
  \label{fig:supp_failure}
\end{figure*}

Although our system significantly improves the metric consistency of
monocular depth, several limitations remain.

First, the method inherits the fundamental weakness of triangulation:
when the camera motion is dominated by rotation or the parallax is too
small, triangulated depths become unreliable. In such cases we fall
back to flow–based warping, which stabilizes the scale but can still be
affected by flow errors.

Second, our refinement relies on the structural continuity of the
relative depth map. When the relative–depth predictor becomes unstable
(e.g., in low-light scenes, overexposed regions, or very distant
surfaces), the propagated scale may cause mild flickering.

Third, in thermal (LWIR) imaging, optical flow estimation can degrade significantly due to low texture, thermal crossover, or sensor bloom, even when FieldScale preprocessing is applied. In such cases, both flow-based scale estimation and pose recovery become unreliable, which can temporarily destabilize the scale tracking module.

Fourth, the approach assumes a predominantly rigid scene. If a large
portion of the frame is covered by independently moving objects, the
flow and triangulation constraints become inconsistent, and the fusion
module may temporarily lose stability.

Finally, extremely distant regions (above 50–80\,m) remain challenging
because both triangulation and flow provide weak geometric cues. This
limitation is shared across all monocular metric–depth approaches.

\end{document}